%% file: main.tex
\documentclass[letterpaper, 10 pt, conference]{ieeeconf}  

\IEEEoverridecommandlockouts                              
\usepackage{amsmath,amssymb,amsfonts}
\usepackage{bm,bbm,siunitx,hyperref,graphicx}
\usepackage{float}	
\usepackage{xcolor}
\usepackage{setspace}
\usepackage{bm}

\let\labelindent\relax

\usepackage{enumitem}
\usepackage{makecell}
\usepackage{cellspace}
\usepackage{cite}
\usepackage{soul}
\usepackage[normalem]{ulem}
\usepackage{amsmath}
\usepackage{cancel}
\soulregister\cite7
\soulregister\ref7
\soulregister\pageref7
\soulregister\label7
\soulregister\item7
\setstcolor{red}

\makeatletter
\newcommand{\hll}[1]{%
    \setbox\@tempboxa\hbox{#1}%
    \ifdim\wd\@tempboxa>\linewidth
    \noindent
    \colorbox{yellow}{%
        \parbox{\dimexpr\linewidth-2\fboxsep}{#1}%
    }%
    \else
    \colorbox{yellow}{#1}%
    \fi}
\makeatother

\usepackage[most]{tcolorbox}
\usetikzlibrary{patterns}
\pgfdeclarepatternformonly{mystrikeout}{\pgfqpoint{-1pt}{-1pt}}{\pgfqpoint{11pt}{11pt}}{\pgfqpoint{10pt}{10pt}}%
{
  \pgfsetlinewidth{0.4pt}
  \pgfpathmoveto{\pgfqpoint{0pt}{0pt}}
  \pgfpathlineto{\pgfqpoint{10.1pt}{10.1pt}}
  \pgfusepath{stroke}
}
\newtcolorbox{tcbstrikeout}{breakable,
 enhanced jigsaw,
 opacityback=0,
 parbox=false,
 boxrule=0mm,
 top=0mm,bottom=0pt,left=0pt,right=0pt,
 boxsep=0pt,
 frame hidden,
 finish={\fill[pattern=mystrikeout] (frame.north west) rectangle (frame.south east);}
}

\usepackage[skip=10pt,font=small]{caption}
\usepackage{balance}   

\title{\LARGE \bf
A flow disturbance estimation and rejection strategy for multirotors with round-trip trajectories
}

\author{
Jaeseung Byun$^{1}$, Simo A. M\"akiharju$^{2}$, and Mark W. Mueller$^{1}$
\thanks{The authors are with $^{1}$High Performance Robotics Lab and $^{2}$FLOW Lab, Mech. Eng. Dept., University of California, Berkeley, CA 94720, USA.
{\tt\small \{jaeseungbyun, makiharju, mwm\}@berkeley.edu}} }
\begin{document}
 
\maketitle

\begin{abstract}
This paper presents a round-trip strategy of multirotors subject to unknown flow disturbances. {During the outbound flight, the vehicle immediately utilizes the wind disturbance estimations in feedback control, as an attempt to reduce the tracking error. During this phase, the disturbance estimations with respect to the position are also recorded for future use. For the return flight, the disturbances previously collected are then routed through a feedforward controller.} {The major assumption here is that the disturbances may vary over space, but not over time during the same mission.} We demonstrate the effectiveness of this feedforward strategy via {experiments with two different types of wind flows; a simple jet flow and a more complex flow.}  {To use as a baseline case, a cascaded PD controller with an additional feedback loop for disturbance estimation} was employed for outbound flights. To display our contributions regarding the additional feedforward approach, {an additional feedforward correction term} obtained via prerecorded data was integrated for the return flight. Compared to the baseline controller, the feedforward controller was observed to produce 43\% {less RMSE position error} at a vehicle ground velocity of \SI{1}{m/s} with \SI{6}{m/s} of environmental wind velocity. This feedforward approach also produced 14\% {less RMSE position error} for the complex flows as well.\footnote{Experimental validation video can be found here:\\ \href{https://youtu.be/lHJLIt3Ul5U}{https://youtu.be/lHJLIt3Ul5U}}
\end{abstract}

\input{1introduction.tex}
\input{2estimationModel.tex}
\input{4algorithm.tex}

\input{5validation.tex}

\input{6conclusion.tex}

\section*{Acknowledgment}
We acknowledge financial support from NAVER LABS, Berkeley Deep Drive consortium, and Code42 Air.
The experimental testbed at the HiPeRLab is the result of contributions of many people, a full list of which can be found at \url{hiperlab.berkeley.edu/members/}.

The authors wish to acknowledge Christian Castaneda Cuella for helping with manufacturing of the nozzle. 

\bibliographystyle{IEEEtran}
\bibliography{main}
\balance

\end{document}

%% file: 1introduction.tex
\section{Introduction}\label{sec:Intro}
Multirotors are {becoming more} widely used in urban areas for various tasks, such as photography \cite{germen2016alternative}, usage of these vehicles has been extending to entertainment {industry} \cite{schollig2010synchronizing}, \cite{10.1145/3196709.3196798}, cleaning windows and solar cells \cite{hassanalian2017classifications}, and inspection \cite{mota2014expanding}. Among many other possible usages, delivery is one of the most practical applications, {such as {drone logistics}} \cite{amazon}, {firefighting (delivering fire extinguishers)} \cite{aydin2019use}, {and blood delivery} \cite{ackerman2019blood}.  {It is projected that 1.4 billion express packages will be delivered by drones by 2030} \cite{nasa2019Report}, {and the total revenue from parcel delivery will be more than 200 billion USD per year in the US alone.}\cite{frachtenberg2019practical}

 Although multirotors have properties that make them suitable for delivery vehicles, such as high agility and controllability, unknown wind disturbances can cause challenges. {For example, disturbances caused by environmental air flows will induce positional and/or altitude errors of the flight trajectory which may cause the multirotor to collide with the surrounding obstacle. {The resulting collision can cause damage to the vehicle or delay the mission.}

Identification and rejection of external disturbances, including wind disturbances, has been extensively studied. {A disturbance rejection using} IMU based estimation is the most {commonly used method}. An integral sliding mode control with a purely IMU-based disturbance observer was shown in \cite{tomic2014evaluation}, and was verified through a series of simulations and experiments. {Although the sliding mode controller showed the best performance compared to PD based disturbance observer}, there was no significant improvement compared to the experimental results that were obtained using a PID controller. This estimation method was extended in \cite{tomic2017external}, {where the forces are estimated from the accelerometer and the torques are estimated using rate gyro sensor data}. This method was designed to filter the undesired noise arising from the numerical differentiation of the rate gyro data. High pass filtered force estimates were {applied to detect collisions with surrounding obstacles.}

{Another feasible disturbance rejection method is to apply optimal control approaches such as model predictive control (MPC). For example,} a nonlinear MPC was implemented by \cite{hentzen2019disturbance}. For the disturbance estimation, an extended Kalman filter (EKF) and an unscented Kalman filter were both applied. {Compared to the baseline controller (PID),} the result showed exceptional rejection performance in terms of {reducing position errors}. Most recently, MPC based trajectory combined with a sliding mode controller was fully implemented and tested in dynamic landing of a quadcopter by \cite{9197081}. The MPC algorithm was performed offboard, {which is similar to the implementation of }\cite{hentzen2019disturbance}.

These control methods are devised as general-purpose approaches, and do not consider the specific flight pattern of a vehicle. Considering a vast number of future delivery missions will be round trips \cite{nasa2019Report}, a specialized control strategy for such round-trip situations can be considered practical. While wind disturbance data can be easily collected during the outbound flight, previous disturbance rejection methods fail to extract and utilize this information during the return flight. 
Thus, we propose a wind disturbance rejection strategy that can improve the flight trajectory of the return trip. {The major assumption in this study is that the disturbance flow is relatively steady and the flow only varies over space within the timeframe of interest.} {Although this assumption can limit the applicability of this rejection strategy, historical weather forecast data show that, such environmental wind-flow patterns exist} \cite{SFweatherForecast}. {Once} the vehicle records the wind disturbances along its planned route {during the outbound flight}, the collected data is fed into the feedforward controller {during the return flight}, which allows the vehicle to move {faster than the vehicle speed of the outbound flight} while still {maintaining smaller positional errors}. {Limitations of this assumption will further be discussed in \ref{discussion} and \ref{sec:conclusion}.}

%% file: 2estimationModel.tex
\begin{figure}
\hfill\includegraphics[width=\columnwidth]{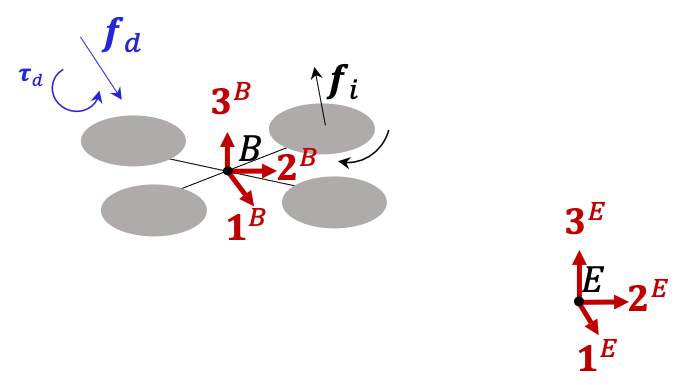}
\caption{A quadcopter experiencing flow disturbance force $\bm{f}_{d}$ and torque $\bm{\tau}_{d}$; body frame and inertial coordinate frames}
\label{fig:FBD}
\end{figure}

\begin{figure}
\hfill\includegraphics[width=\columnwidth]{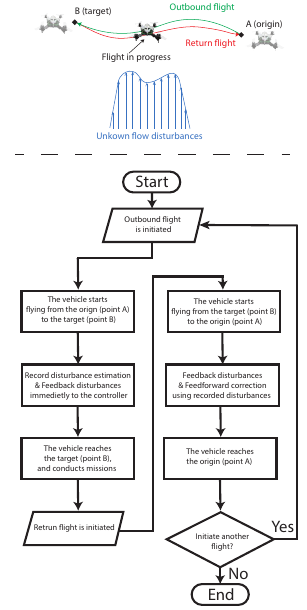}
\caption{Schematic of the round-trip trajectory under unknown flow disturbances (top) and flow chart of the disturbance rejection strategy for round-trip flights (bottom)}
\label{fig:flowChart}
\end{figure}
\section{Disturbance estimation model} \label{sec:DistModel}
In this section, we introduce a disturbance estimation model for external forces and torque for a quadrotor. {A disturbance observer is utilized for the disturbance estimation, similar to the method used in \cite{tomic2017external}.} We use the earth local frame as an inertial frame, {while the body frame is aligned with the vehicle}. 
\subsection{Disturbance force estimation model}\label{sec:FoceEstimation}
 {We assume that the total force is compose of the disturbance force, the sum of propeller thrusts, and the weight. Applying Euler's first law,}
 \begin{equation}\label{eq:distForce}
m\bm{a}=\sum_{i=1}^{4}\bm{f}_{i}+\bm{f}_{d}+m\bm{g}
\end{equation}
where $\bm{f}_{i}$ is the force produced by each propeller, $\bm{f}_{d}$ is the disturbance force, $m$ is mass of the body, $\bm{a}$ is the linear acceleration, and $\bm{g}$ is gravitational acceleration. Here we assume that the disturbance forces act on the mass center of the body. Accelerometer readings $\bm{\alpha}$ can be represented as a difference between the linear acceleration and gravity, \begin{equation}
\bm{R}\bm{\alpha}=\bm{a}-\bm{g}
\end{equation} 
where $\bm{R}$ denotes a rotation matrix which converts the body frame to the earth frame. Then (\ref{eq:distForce}) becomes
\begin{equation}
\bm{f}_{d}=m\bm{R}\bm{\alpha}-\sum_{i=1}^{4}\bm{f}_{i}\label{eq:distForceModel}
\end{equation}
Assuming the motor force produced from the command thrust is the same as the true force produced by the propellers (i.e., secondary effects such as blade flapping \cite{mahony2012multirotor} are not considered), then (\ref{eq:distForceModel}) becomes

\begin{equation}
\bm{f}_{d}=m{\bm{R}}\bm{\alpha}-c_{\Sigma}\bm{3}^B
\end{equation} 
where $c_{\Sigma}$ denotes the sum of the motor forces {and $\bm{3}^B$ is a unit vector component in the body frame.}




\subsection{Disturbance torque estimation model}\label{sec:TorqueEstimation}
The total torque acting on the quadcopter ${\bm{\tau}}_{tot}$ {can be represented in terms of angular acceleration and angular velocity by Euler's second law,}
\begin{equation}{\bm{\tau}}_{tot}=\frac{d}{dt}\left(\bm{J}\bm{\omega}\right)+\bm{\omega}\times(\bm{J}\bm{\omega})\label{eq:EulerLaw}\end{equation}

where $\bm{J}$ represents mass moment of inertia of the body, and $\bm{\omega}$ is the angular velocity. Total torque is composed of the disturbance torque ${\bm{\tau}}_{d}$ and the torque produced from the collection of propellers ${\bm{\tau}}_{p}$. Then equation (\ref{eq:EulerLaw}) becomes

\begin{equation}
{\bm{\tau}}_{d}=\bm{J}\frac{d}{dt}\bm{\omega}+\bm{\omega}\times(\bm{J}\bm{\omega})-{\bm{\tau}}_{p}\label{eq:true_torque}
\end{equation}


%% file: 4algorithm.tex
\section{Round-trip disturbance rejection strategy}\label{algorithm}


In this section we lay out our approach for our disturbance rejection strategy specialized for round trips. In Fig. \ref{fig:flowChart}, we depict a multirotor during a round-trip mission. Consider the scenario where the vehicle is making a round trip between the origin A and a target point B, with unknown wind flow along the expected trajectory. We further divide this round-trip flight into 4 different stages:
\begin{enumerate}
    \item {After starting the outbound flight from point A, we record the flow disturbance force and torque estimated via the IMU sensor until we reach the destination (point B). The according position of the vehicle is also stored alongside these data.
}\label{outbound}
    \item {The vehicle performs its planned mission when after arriving at point B.}
    \item {The vehicle now returns to point A, but during this trip we feedforward the recorded disturbance force and torque. These additional feedforward inputs allow us to operate the vehicle at greatly elevated speeds with stability.
}
    \item {The vehicle safely returns to the origin, and may repeat step 1 when assigned with a new mission.}
\end{enumerate}
\subsection{Controller and disturbance estimation}\label{ctrlEst}
\begin{figure}
    \centering
    \includegraphics[width=\columnwidth]{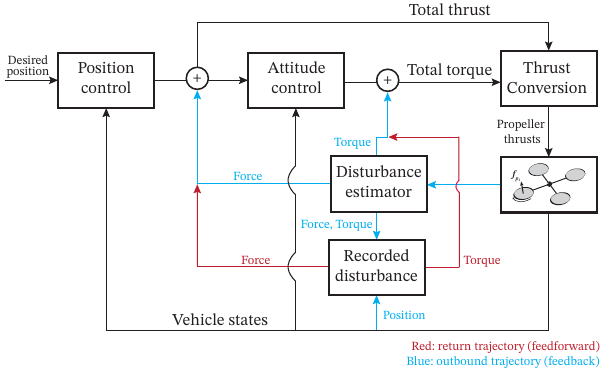}
    \caption{Block diagram of the controller in the round-trip strategy. Red arrow is used for the return flight}
    \label{fig:Controller}
\end{figure}
We consider two different types of controllers. The controller used for the outbound flight is a cascaded PD position and attitude controller which runs offboard. (Fig. \ref{fig:Controller}) The controller used for the outbound flight employs a feedback-based disturbance rejection method;
\begin{itemize}
\item The controller prioritizes minimizing the thrust direction error via a nonlinear attitude controller. \cite{mueller2018}
\item The controller considers the estimated disturbance forces and torques and includes that in the feedback loop. This allows smaller tracking errors under disturbances compared to controllers without this feature.
\end{itemize}
The outbound flight controller is where further improvements are made via the feedforward strategy. In addition to the feedback loop, the prerecorded disturbance data is fed in as an input. 
{The disturbance force can be estimated as,}
\begin{equation}
{\bm{\hat{f}}}_{d}=\mathcal{F}\left(m\bm{R}\tilde{\bm{\alpha}}-c_{\Sigma}\bm{3}^{B}\right)\label{FilterEstimation}
\end{equation}
 Where $\mathcal{F}\left(\cdot\right)$ is a low-pass filter operation, and ${\bm{\hat{f}}}_{d}$ is the estimated disturbance force from the accelerometer measurement $\tilde{\bm{\alpha}}$.  
 
 The disturbance torque estimate ${\bm{\hat{\tau}}}_{d}$ can be written in terms of torque from propellers ${\bm{{\tau}}}_{p}$, angular velocity measurement using rate gyro sensor $\tilde{\bm{\omega}}$, and angular acceleration estimate {$\widehat{\frac{d}{dt}\bm{\omega}}$. The angular acceleration is estimated from the numerical differentiation of the rate-gyro measurements, where a low-pass filter was applied to.} Applying the model given in (\ref{eq:true_torque}),
\begin{equation}
{\bm{\hat{\tau}}}_{d}=\mathcal{F}\left(\bm{J}\widehat{\frac{d}{dt}\bm{\omega}}+\bm{\tilde{\omega}}\times(\bm{J}\bm{\tilde{\omega}})-{\bm{{\tau}}}_{p}\right)
\end{equation}
\subsection{Outbound flight}
\label{OutboundFlight}\label{outboundFlight}
{When the vehicle is making the outbound flight, the disturbance estimation developed from \ref{ctrlEst} is continuously computed offboard. The current disturbance estimation is also utilized as an additional component of the feedback control. (Fig. \ref{fig:Controller})} {The updated force estimate of the outbound flight} ${\bm{\hat{f}}}_{{d},\text{out}}[k]$ {simply becomes,}
\begin{equation}
{\bm{\hat{f}}_{{d},\text{out}}}\left[k\right]={\bm{\hat{f}}_{{d},\text{out}}}\left[k-1\right]
\label{estimate1}
\end{equation}

 It is important to note that the outbound rejection algorithm must provide quality estimates so that minimum position error can be ensured. Thus, we limit the speed of the vehicle for the most accurate trajectory.  At the same time, we record the estimated disturbance torque and force as a function of the current position. {The position is estimated offboard; we note that the sampling rate of the disturbance records is smaller than that of the onboard data collection. In our experience, downsampling the higher frequency data streams provided enough fidelity for our application.}
 

\subsection{Return flight}\label{returnFlight}
The estimation method is the same as the outbound flight, except that the trajectory is now reversed. The major difference is that we now conduct feedforward control with the recorded data, with respect to the current position. The closest distance between the current position and the position of the outbound trip trajectory is found from {brute-force searching algorithm}, and recorded disturbance force (${\bm{\hat{f}}_{{d},\text{rec}}}$) and torque at that particular position are extracted for rejecting disturbances. 

The recorded disturbances are used to improve our position estimator. The updated force estimate of the return flight ${\bm{\hat{f}}}_{{d},\text{return}}[k]$ is,
\begin{equation}
{\bm{\hat{f}}_{{d},\text{return}}}\left[k\right]=\mathcal{F}\left({\bm{\hat{f}}_{{d},\text{return}}}\left[k-1\right]-{\bm{\hat{f}}_{{d},\text{rec}}}\left[k\right]\right)+{\bm{\hat{f}}_{{d},\text{rec}}}\left[k\right]
\label{estimate}
\end{equation}
and we can achieve the position update by integration of the acceleration term deduced from the force update.
A notable distinction from the outbound flight is that we now use wind disturbance a-priori. This not only enables us to ensure the vehicle is able to follow an accurate trajectory since disturbance of current time step is available, but also allows the vehicle to return faster with similar positional errors to that of outbound flight.

%% file: 5validation.tex
\section{Experimental Validation}\label{sec:ExpValid}
{Three experiments} were carried out to validate our control strategy. The position error, attitude error, disturbance force, and disturbance torque are shown with respect to the flight time, and noticeable observations from the experimental results will be discussed.
\subsection{Experimental setup}


\begin{table}
\begin{centering}
\caption{Physical parameters of the test vehicle}
\label{table:vehSpecs}
\begin{spacing}{1.1}
\begin{tabular}{c|p{17mm}}
 Prop. diameter
 (\SI{}{\mm}) & \hfil 50.8\tabularnewline
\hline 
 Arm length (\SI{}{\mm}) & \hfil 58.5\tabularnewline
\hline 
 Mass (\SI{}{\gram}) & \hfil 154 \tabularnewline
\hline
 Maximum thrust (\SI{}{\N})  & \hfil 4.6 \tabularnewline
\end{tabular}
\end{spacing}
\par\end{centering}
\end{table}

\begin{figure}
\hfill\includegraphics[width=\columnwidth]{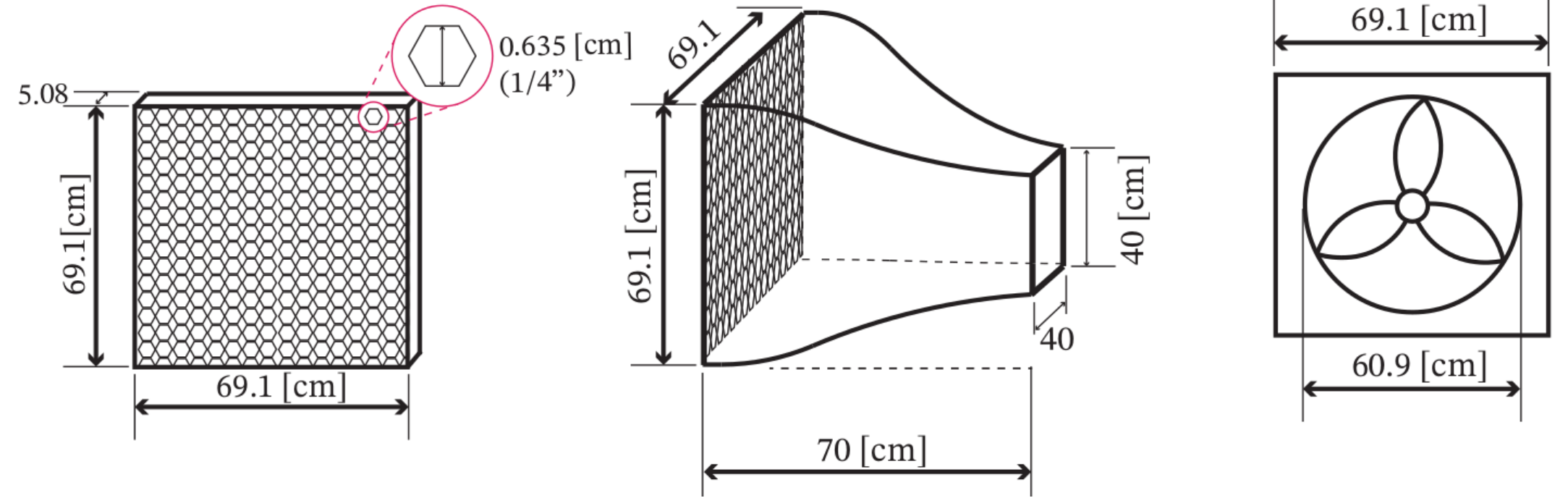}
\caption{Dimensions for the test facility}
\label{fig:nozzle}
\end{figure}

\begin{figure}
\centering
\includegraphics[width=0.35\textwidth]{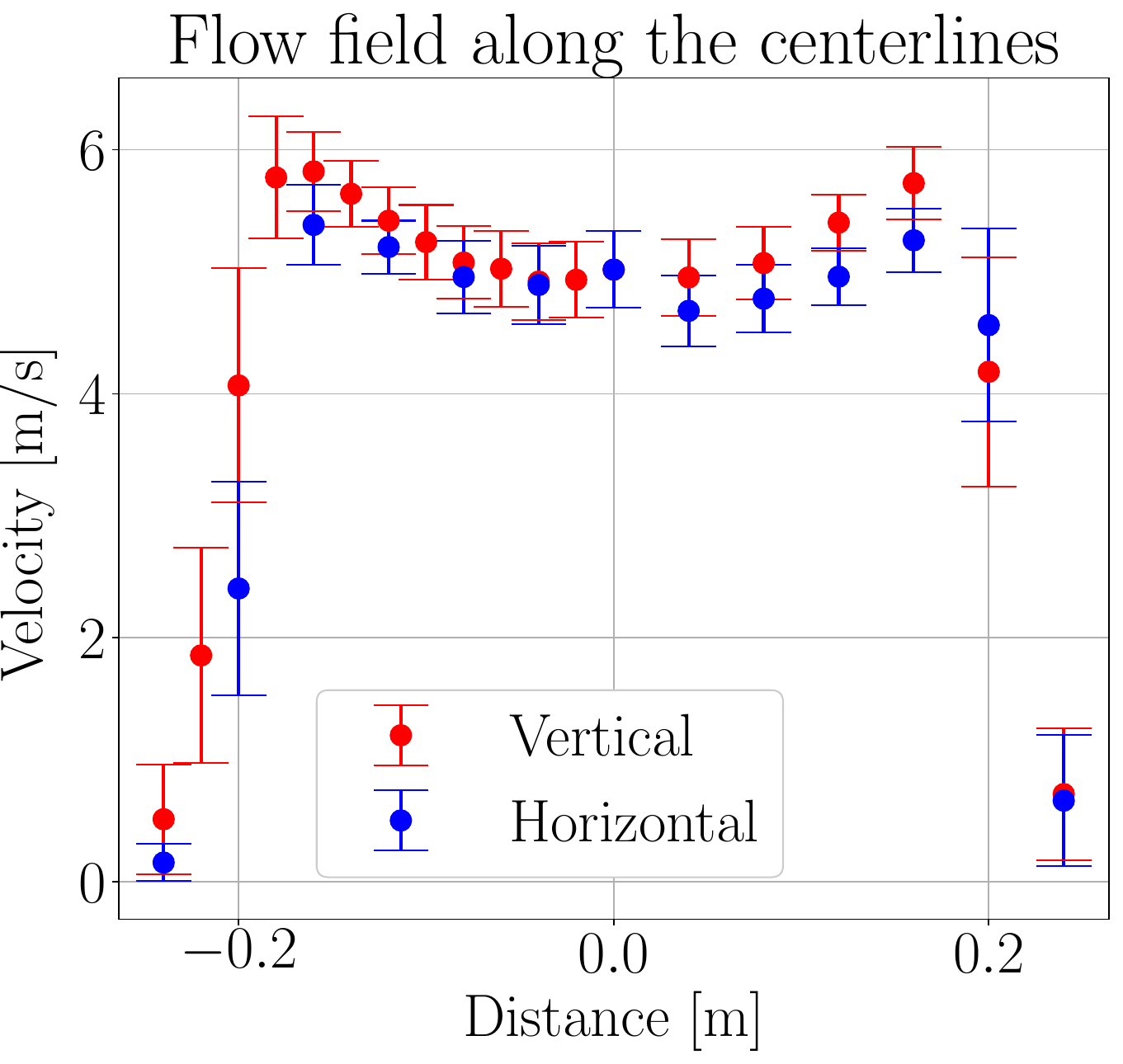}
\caption{Flow field identification, $\SI{30}{cm}$ far from the nozzle exit. Error bar denotes standard deviation of the measured velocity}
\label{fig:flowProfile}
\end{figure}
The quadcopter for our demonstration and its physical parameters are described in Table \ref{table:vehSpecs}. Localization is based on a motion capture system, complemented with the rate gyro measurement. The frequency of the radio command signal is $\SI{50}{Hz}$. {The total thrust is computed offboard. Since we desire to use the disturbance force as a feedforward correction during the return flight, the disturbance force is recorded at a sampling rate of} $\SI{50}{Hz}$ during the outbound flight.{The total torque estimation is computed onboard at $\SI{500}{Hz}$, while the disturbance torque is recorded at $\SI{50}{Hz}$}. Downsampling of the higher frequency stream was done to match the two streams.} The recording time horizon was around \SI{25}{s} for all experiments. The position estimation was also computed so that it was available at a rate of {$\SI{200}{Hz}$}. 

The nozzle used to generate the wind disturbance was capable of generating up to $\SI{6}{m/s}$ of environmental wind velocity at $\SI{30}{cm}$ distance from its nozzle exit. The detailed flow profile is given in Fig. \ref{fig:flowProfile}, and the specific dimensions are shown in Fig. \ref{fig:nozzle}. An accurate measure of the positions shown in the flow field was achieved by the motion-capture system. 

\begin{figure}
\centering\includegraphics[width=0.27\textwidth]{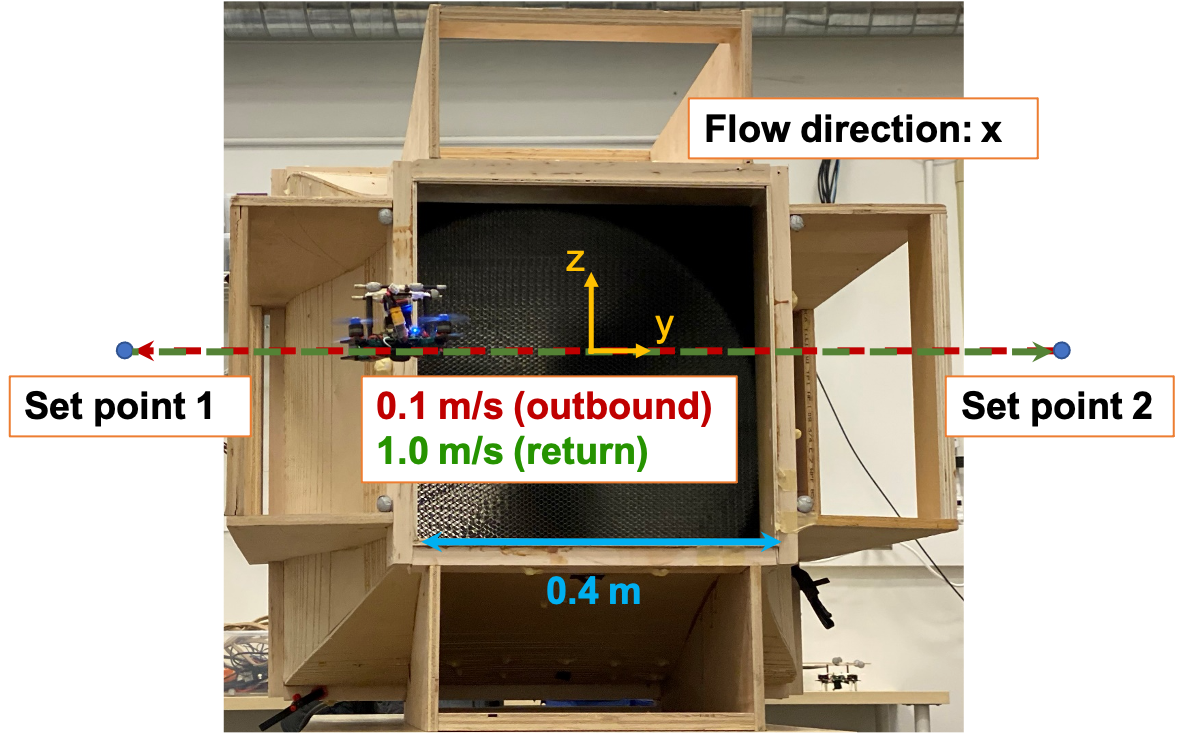} \includegraphics[width=0.20\textwidth]{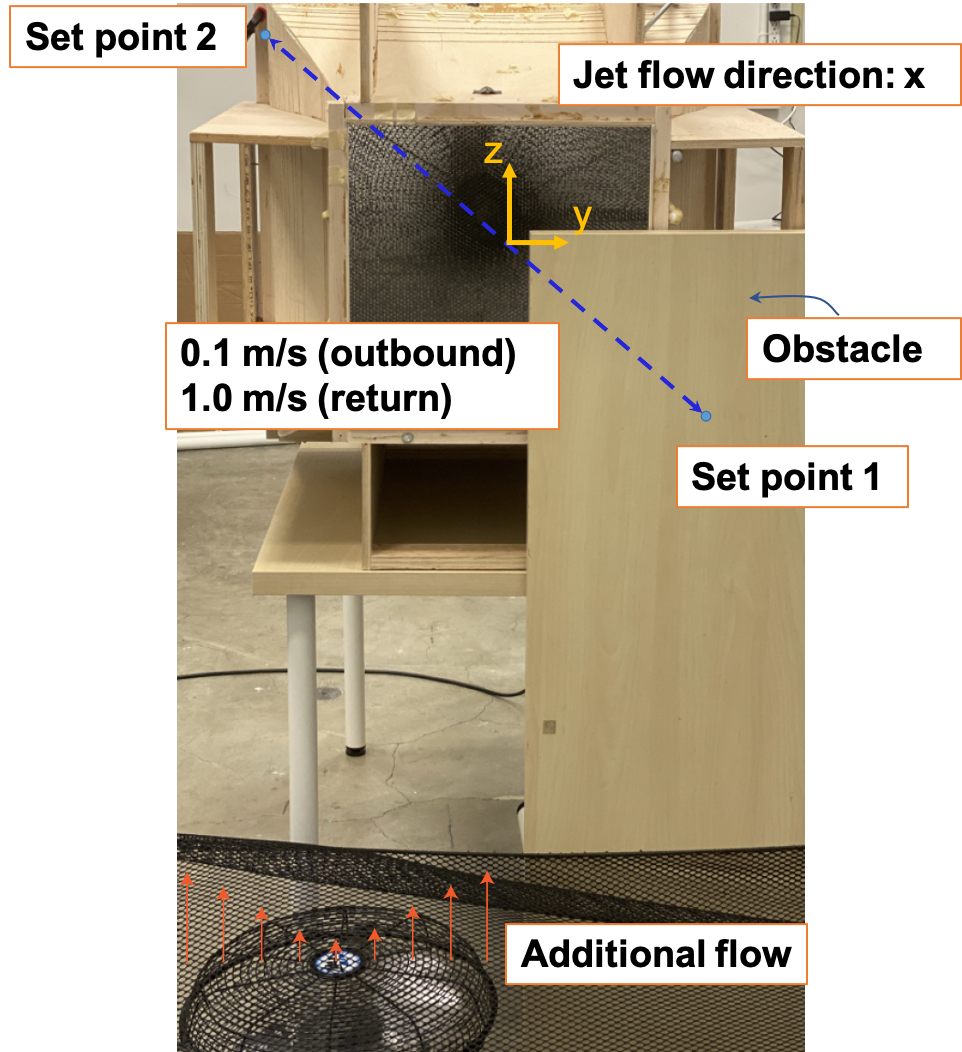}
\caption{Description of straight-line trajectory with jet flow (left) and complex flow field set-up (right)}
\label{fig:repeatingTraj}
\end{figure}

\subsection{{Disturbance rejection performance of the baseline controller during the outbound flight}}
\begin{figure}
\centering\includegraphics[width=\columnwidth]{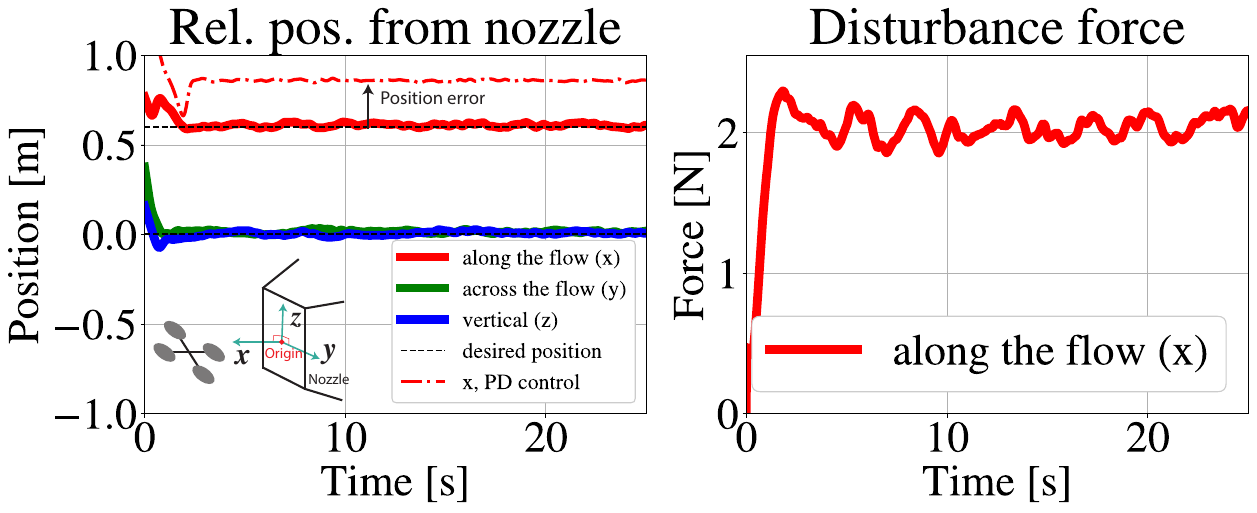}
\caption{Static point experiments, the vehicle at hover under flow disturbance}
\label{fig:static}
\vspace{0.2cm}
\centering\includegraphics[width=\columnwidth]{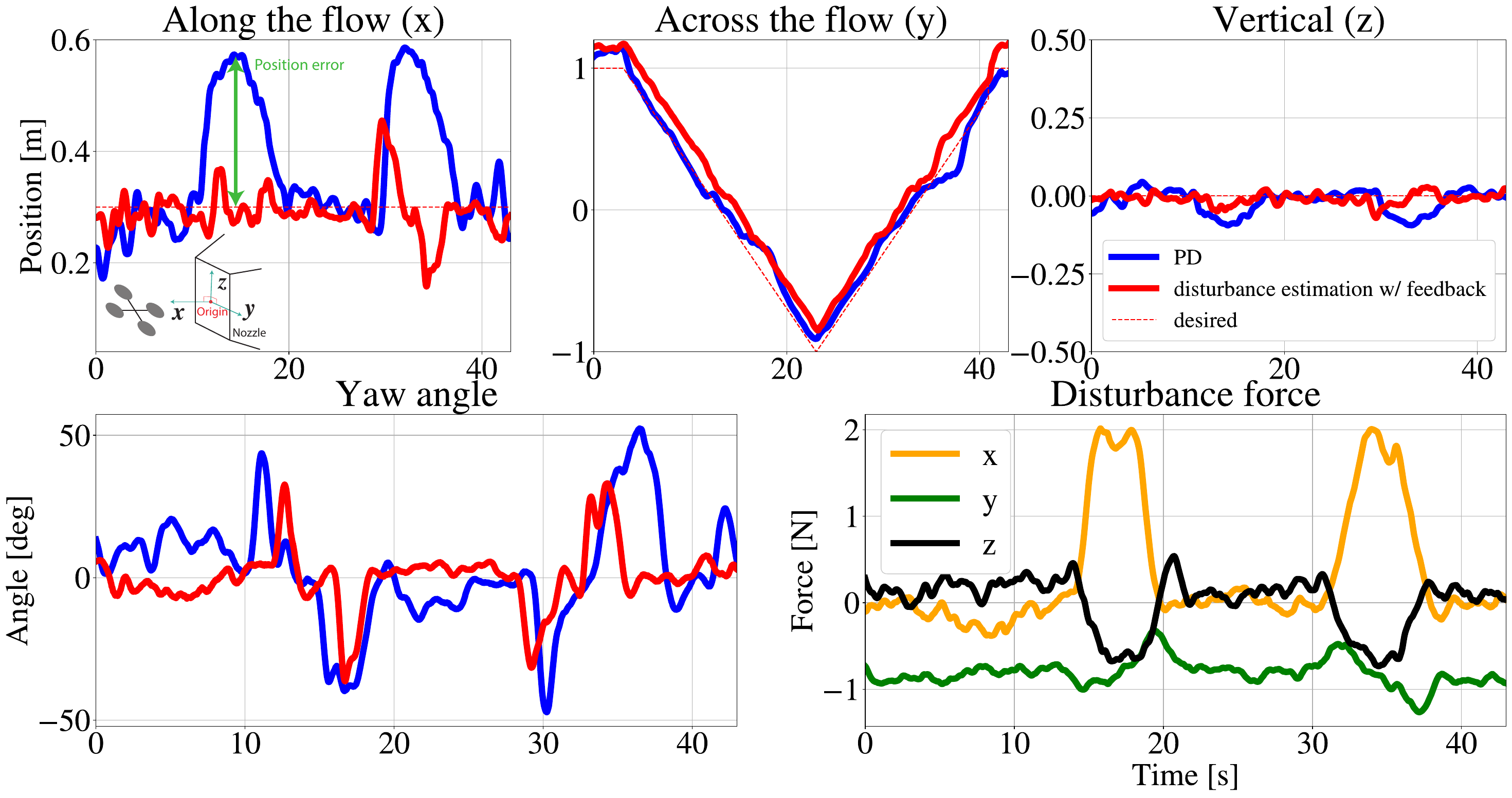}
\caption{Round-trip tracking performance of the baseline controller with flight velocity of \SI{0.1}{m/s}}
\label{fig:repeat}
\end{figure}

{Obtaining accurate position estimates (minimizing tracking errors) from our baseline controller during the outbound flight is an important task for our strategy since this information is further utilized on the return flight. We verify the tracking performance (in terms of RSME position error) of the baseline controller by comparing our baseline controller to a conventional PD controller. We consider the case of hovering} (Fig. \ref{fig:static}) and the case of a round trip between two points where the trajectory of the vehicle is subject to a simple jet ﬂow. The vehicle velocity used was \SI{0.1}{m/s} (Fig. \ref{fig:repeat}). {The tracking performance of the baseline controller will be further discussed in \ref{discussion}.}



\begin{figure*}
\centering\includegraphics[keepaspectratio, width=17.5cm]{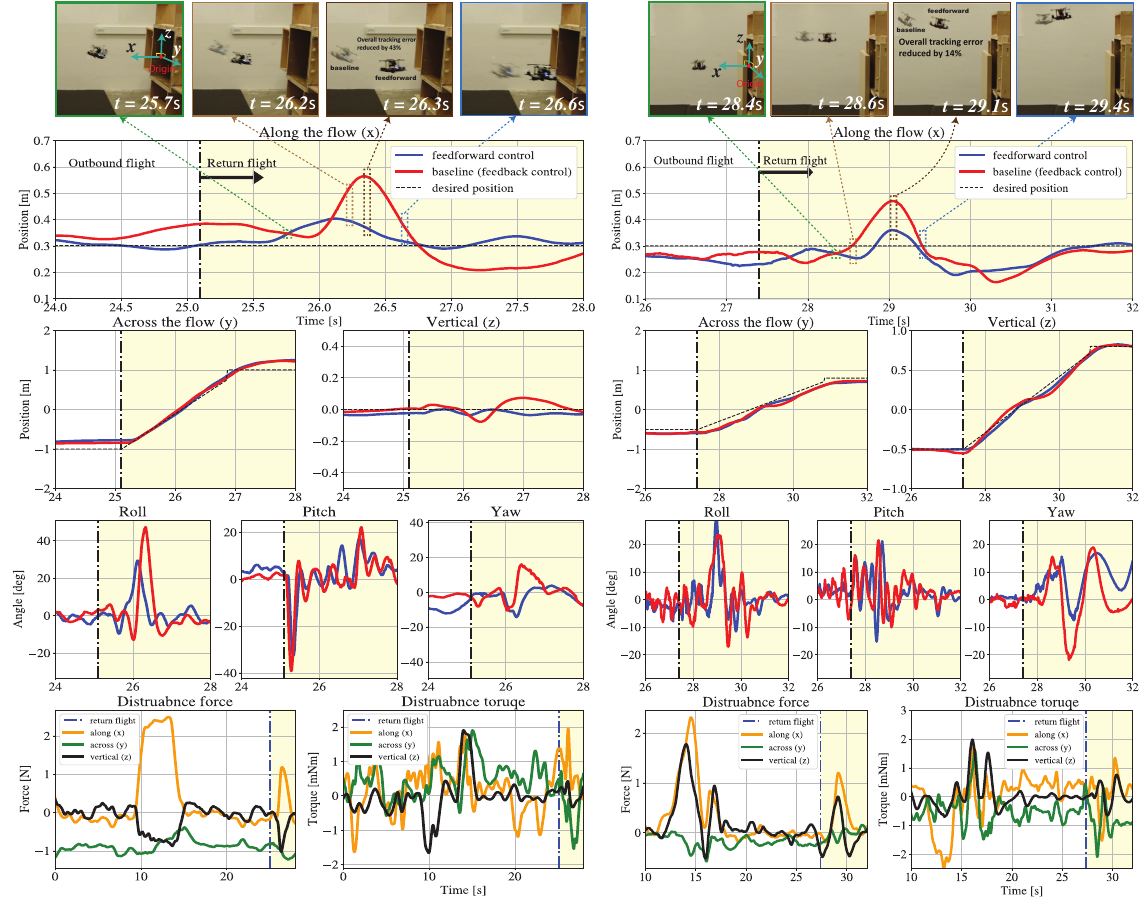}
\caption{Round-trip trajectory experiment with jet flow (left) and complex flow (right)}
\label{fig:roundTrip}
\end{figure*}

\subsection{Round-trip experiments with feedforward correction of recorded disturbance data}
\subsubsection{Trajectory with jet flow disturbance (Fig. \ref{fig:repeatingTraj})}
 Finally, we test the performance of the new disturbance rejection strategy. For the purpose of this performance test, we use two straight-line trajectory flights, while the vehicle is flying back at 10 times larger speed than that of the outbound flight. Here we chose a return flight velocity of \SI{1.0}{m/s}.
\subsubsection{Flight trajectory subject to complex wind flows}
In order to test our strategy against a flow that can be considered less orderly, we use a flow field that was designed as shown in Fig. \ref{fig:repeatingTraj}. A quadrant of the nozzle exit was blocked by an obstacle, and an additional fan was installed on the floor. The produced ﬂow points upwards, and mixes with the ﬂow from the nozzle to create a more complex disturbance.

%

\subsection{Results and discussion}\label{discussion}
\subsubsection{{Tracking performance results of baseline controller} {with vehicle traveling at \SI{0.1}{m/s}}}
{Position and disturbance force estimation of the vehicle subject to a static ﬂow ﬁeld is shown in Fig. \ref{fig:static}. The largest position error was observed with the PD controller along the flow direction (Fig. \ref{fig:repeat}). This behavior is expected since the disturbance force from the ﬂow is mainly expected to act along the ﬂow direction.}

The presence of oncoming ﬂow also affects the vertical position of the vehicle. This is because of the positive drag generated by the positive pitch angle of the vehicle flying forwards \cite{mahony2012multirotor}. Significant yawing motion was observed near the boundary of the ﬂow. This is due to the large velocity gradient at the boundary increasing the drag force only for half of the propeller, producing additional yaw torque.

{We observe that the baseline controller (PD controller with additional disturbance estimation feedback loop) reduces the position errors for the conditions used in our experiment} (Fig. \ref{fig:repeat}). {This implies that the position estimations can be considered accurate enough, thus these recordings may be utilized during the return flight, along with the disturbance force estimates.}

\subsubsection{{Tracking performance of the feedforward controller on the return flight}}
Position, attitude, and complete disturbance estimation of the round-trip experiment with a straight-line trajectory is given in Fig. \ref{fig:roundTrip}. The yellow shaded area denotes the return trajectory with a velocity (\SI{1}{m/s}) ten times higher than that of the outbound trip (\SI{0.1}{m/s}). {The RSME positional error was improved noticeably (error reduced by 43\%). }
Overall, we observe that despite the introduction of the complex ﬂow ﬁeld for the round-trip scenario (described in Fig. \ref{fig:repeatingTraj}), {the RSME position error remained small (error reduced by 14\%).} We also notice that the case where the recorded disturbance data was not utilized, performed the worst. This indicates that the recording strategy can provide resilient counteraction for the given ﬂow ﬁeld, even with complex flow disturbances.

An interesting observation that can be made is that the disturbance force was underestimated by half during the return flight. This can be due to the fact that faster operation speeds compared to that of the outbound flight hinders the capability to accurately capture the disturbance force. This underestimation of disturbances ultimately led to increased positional error when solely using the baseline controller. On the other hand, the feedforward controller continued to show smaller positional errors even in such situations where accurate disturbance readings were not available.

\subsubsection{{Limitations of `relatively steady flow' assumption}}
{The major assumption we made in this paper is that the environmental ﬂow is expected to stay relatively steady, and the characteristic of the ﬂow mostly varies spatially. In reality, this assumption might not always be the case. Once the ﬂow becomes unsteady, the recorded disturbance data can no longer be considered valid, and will result in even larger position errors under feedforward control.} {One possible solution is to frequently check the weather condition via weather forecast information or utilizing other available weather prediction data near the region of interest. We then decide whether we wish to continue to employ the feedforward control strategy. In the worst-case scenario, we may always resort back to the baseline controller (feedback-based controller) with a reduced vehicle speed. Our feedforward rejection strategy may be utilized as soon as it is determined that the weather condition meets our assumption, which will allow a significant reduction of mission times compared to solely relying on the baseline controller.}

%% file: 6conclusion.tex
\section{Conclusions and future work} \label{sec:conclusion}
In this paper, a simple and novel disturbance rejection algorithm specializing in round trips for multirotors has been introduced. The key idea is to record the disturbance estimations as a function of the position for the outbound flight, and use the recorded data as an input for a feedforward control strategy.

Experiments that cover several round-trip scenarios were conducted with our suggested algorithm. Overall, during the return flight, RSME position error was greatly reduced when the recorded disturbance estimation obtained from the outbound flight was fed into our control algorithm as a feedforward input. {Experimental results demonstrated that} the vehicle being operated at \SI{1}{m/s} with relatively steady state surrounding flows at around \SI{6}{m/s}, RSME position error was reduced by 43\% compared to the baseline feedback-based controller. With more complex wind flows, specifically for the case depicted in the previous section, the algorithm still proved useful, reducing tracking errors by 14\%.

Investigation of improving this strategy for more unsteady environments is under consideration for future work. The main consideration will be to further develop a feedforward control-based rejection strategy to deal with {greater degree of} transitory environments. An MPC-like controller may also be employed in this case. {Unsteady wind gust model such as Dryden wind turbulence model can also be implemented.}

{Future application includes round trips with more challenging environments, such as a round-trip scenario with near-boundaries - where the ground or ceiling effect becomes more significant.} Also, the motion capture-based localization system used in our experiments currently prevents us from performing large-scale experiments. Preferably, an onboard position localization method will allow us to investigate such disturbance rejection strategies in an outdoor environment. {More advanced estimation methods including extended Kalman filter or a dynamic observer may further improve the estimation error of the outdoor flight trajectory.}

%% file: main.bbl
\begin{thebibliography}{10}
\providecommand{\url}[1]{#1}
\csname url@samestyle\endcsname
\providecommand{\newblock}{\relax}
\providecommand{\bibinfo}[2]{#2}
\providecommand{\BIBentrySTDinterwordspacing}{\spaceskip=0pt\relax}
\providecommand{\BIBentryALTinterwordstretchfactor}{4}
\providecommand{\BIBentryALTinterwordspacing}{\spaceskip=\fontdimen2\font plus
\BIBentryALTinterwordstretchfactor\fontdimen3\font minus
  \fontdimen4\font\relax}
\providecommand{\BIBforeignlanguage}[2]{{%
\expandafter\ifx\csname l@#1\endcsname\relax
\typeout{** WARNING: IEEEtran.bst: No hyphenation pattern has been}%
\typeout{** loaded for the language `#1'. Using the pattern for}%
\typeout{** the default language instead.}%
\else
\language=\csname l@#1\endcsname
\fi
#2}}
\providecommand{\BIBdecl}{\relax}
\BIBdecl

\bibitem{germen2016alternative}
M.~Germen, ``Alternative cityscape visualisation: drone shooting as a new
  dimension in urban photography,'' \emph{Electronic visualisation and the
  arts}, pp. 150--157, 2016.

\bibitem{schollig2010synchronizing}
A.~Sch{\"o}llig, F.~Augugliaro, S.~Lupashin, and R.~D'Andrea, ``Synchronizing
  the motion of a quadrocopter to music,'' in \emph{2010 IEEE International
  Conference on Robotics and Automation}.\hskip 1em plus 0.5em minus
  0.4em\relax IEEE, 2010, pp. 3355--3360.

\bibitem{10.1145/3196709.3196798}
H.~Kim and J.~A. Landay, ``Aeroquake: Drone augmented dance,'' in
  \emph{Proceedings of the 2018 Designing Interactive Systems Conference}, ser.
  DIS ’18.\hskip 1em plus 0.5em minus 0.4em\relax Association for Computing
  Machinery, 2018.

\bibitem{hassanalian2017classifications}
M.~Hassanalian and A.~Abdelkefi, ``Classifications, applications, and design
  challenges of drones: A review,'' \emph{Progress in Aerospace Sciences},
  vol.~91, pp. 99--131, 2017.

\bibitem{mota2014expanding}
R.~L. Mota, L.~F. Felizardo, E.~H. Shiguemori, A.~C. Ramos, and F.~Mora-Camino,
  ``Expanding small uav capabilities with ann: a case study for urban areas
  inspection,'' \emph{British Journal of Applied Science \& Technology},
  vol.~4, no.~2, p. 387, 2014.

\bibitem{amazon}
``{Amazon Prime Air},''
  \url{https://www.amazon.com/Amazon-Prime-Air/b?ie=UTF8&node=8037720011},
  2021, [Online; accessed 22-February-2021].

\bibitem{aydin2019use}
B.~Aydin, E.~Selvi, J.~Tao, and M.~J. Starek, ``Use of fire-extinguishing balls
  for a conceptual system of drone-assisted wildfire fighting,'' \emph{Drones},
  vol.~3, no.~1, p.~17, 2019.

\bibitem{ackerman2019blood}
E.~Ackerman and M.~Koziol, ``The blood is here: Zipline's medical delivery
  drones are changing the game in rwanda,'' \emph{IEEE Spectrum}, vol.~56,
  no.~5, pp. 24--31, 2019.

\bibitem{nasa2019Report}
\BIBentryALTinterwordspacing
S.~Hasan, ``Urban air mobility (uam) market study,'' The National Aeronautics
  and Space Administration, Tech. Rep. HQ-E-DAA-TN70296, June 2019. [Online].
  Available: \url{https://ntrs.nasa.gov/citations/20190026762}
\BIBentrySTDinterwordspacing

\bibitem{frachtenberg2019practical}
E.~Frachtenberg, ``Practical drone delivery,'' \emph{Computer}, vol.~52,
  no.~12, pp. 53--57, 2019.

\bibitem{tomic2014evaluation}
T.~Tomi{\'c}, ``Evaluation of acceleration-based disturbance observation for
  multicopter control,'' in \emph{2014 European Control Conference
  (ECC)}.\hskip 1em plus 0.5em minus 0.4em\relax IEEE, 2014, pp. 2937--2944.

\bibitem{tomic2017external}
T.~Tomi{\'c}, C.~Ott, and S.~Haddadin, ``External wrench estimation, collision
  detection, and reflex reaction for flying robots,'' \emph{IEEE Transactions
  on Robotics}, vol.~33, no.~6, pp. 1467--1482, 2017.

\bibitem{hentzen2019disturbance}
D.~{Hentzen}, T.~{Stastny}, R.~{Siegwart}, and R.~{Brockers}, ``Disturbance
  estimation and rejection for high-precision multirotor position control,'' in
  \emph{2019 IEEE/RSJ International Conference on Intelligent Robots and
  Systems (IROS)}, 2019, pp. 2797--2804.

\bibitem{9197081}
A.~{Paris}, B.~T. {Lopez}, and J.~P. {How}, ``Dynamic landing of an autonomous
  quadrotor on a moving platform in turbulent wind conditions,'' in \emph{2020
  IEEE International Conference on Robotics and Automation (ICRA)}, 2020, pp.
  9577--9583.

\bibitem{SFweatherForecast}
``{National Weather Service Forcast Office, San Francisco Bay Area/Monterey},''
  \url{https://forecast.weather.gov/MapClick.php?lat=37.8717&lon=-122.272&unit=0&lg=english&FcstType=graphical},
  2021, [Online; accessed 21-January-2021].

\bibitem{mahony2012multirotor}
R.~Mahony, V.~Kumar, and P.~Corke, ``Multirotor aerial vehicles: Modeling,
  estimation, and control of quadrotor,'' \emph{IEEE Robotics and Automation
  magazine}, vol.~19, no.~3, pp. 20--32, 2012.

\bibitem{mueller2018}
M.~W. Mueller, ``Multicopter attitude control for recovery from large
  disturbances,'' \emph{arXiv:1802.09143}, 2018.

\end{thebibliography}
